\documentclass{article}
\usepackage{spconf,amsmath,graphicx}
\usepackage{multirow}
\usepackage{booktabs}
\usepackage{tabularray}
\usepackage{indentfirst}
\usepackage[shortlabels]{enumitem}
\usepackage[hidelinks]{hyperref}

\title{F2T2-HiT: A U-Shaped FFT Transformer and Hierarchical Transformer for Reflection Removal}
\name{Jie Cai, Kangning Yang, Ling Ouyang, Lan Fu, Jiaming Ding, Huiming Sun, Chiu Man Ho, Zibo Meng}
\address{OPPO AI Center}

\begin{document}
%
\maketitle
\begin{abstract}

\textbf{S}ingle \textbf{I}mage \textbf{R}eflection \textbf{R}emoval (SIRR) technique plays a crucial role in image processing by eliminating unwanted reflections from the background. These reflections, often caused by photographs taken through glass surfaces, can significantly degrade image quality. SIRR remains a challenging problem due to the complex and varied reflections encountered in real-world scenarios.
These reflections vary significantly in intensity, shapes, light sources, sizes, and coverage areas across the image, posing challenges for most existing methods to effectively handle all cases.
To address these challenges, this paper introduces a U-shaped \textbf{F}ast \textbf{F}ourier \textbf{T}ransform \textbf{T}ransformer and \textbf{Hi}erarchical \textbf{T}ransformer (F2T2-HiT) architecture, an innovative Transformer-based design for SIRR. 
Our approach uniquely combines Fast Fourier Transform (FFT) Transformer blocks and Hierarchical Transformer blocks within a UNet framework.
The FFT Transformer blocks leverage the global frequency domain information to effectively capture and separate reflection patterns, while the Hierarchical Transformer blocks utilize multi-scale feature extraction to handle reflections of varying sizes and complexities.
Extensive experiments conducted on three publicly available testing datasets demonstrate state-of-the-art performance, validating the effectiveness of our approach.

\end{abstract}

\section{Introduction}
\label{sec:intro}

The SIRR technique aims to decompose a reflection-contaminated input image \( I \) into two distinct components: the reflection layer \( R \) and the reflection-free transmission layer \( T \). The primary objective is to ensure that the generated transmission layer \( T \) closely approximates its corresponding ground truth \( B \), which represents the ideal reflection-free image.

This paper proposes a U-shaped network where the Fast Fourier Transform Transformer (F2T2) block and the Hierarchical Transformer (HiT) block serve as fundamental building blocks for generic reflection removal. F2T2-HiT is built upon the elegant architecture of NAFNet~\cite{chen2022simple}, where the convolutional layers are replaced with Transformer blocks while retaining the overall hierarchical encoder-decoder structure and skip connections. Specifically, we propose two core designs to make F2T2-HiT suitable for reflection removal tasks. 

First, we note that a significant limitation of most current SIRR architectures is their limited effective receptive field. 
To tackle this issue, we introduce an FFT-based Transformer block, which acts as an efficient and effective foundational component for global feature extraction. This block allows the model to learn long-range dependencies and capture spatial relationships across the entire input. A large effective receptive field is critical for comprehending an image's global structure, a key requirement for reflection removal.
Moreover, for large reflection areas, even an expanded receptive field achieved through larger or dilated convolutions may still miss the information needed to produce a high-quality reflection-free image.
To overcome this limitation, we leverage Fast Fourier Convolutions (FFCs)~\cite{chi2020fast}, which provide a larger receptive field that spans the entire image even in the early layers of the network. This property of FFCs improves both perceptual quality and parameter efficiency of the network.

Second, we note that most existing SIRR architectures struggle to effectively capture both local details and global context across multiple scales.
To tackle this, we introduce a hierarchical window-partition Transformer block, which processes images through multi-scale window partitions. This design enables the block to capture fine-grained local details while maintaining global context by dividing the input into varying window sizes. 
By integrating multi-scale attention mechanisms, our method achieves a more precise understanding of reflection-transmission relationships while preserving computational efficiency through localized attention. Crucially, this design delivers superior reflection removal performance without significant computational cost.

In summary, our key contributions include:
\begin{itemize}

    \item We propose a novel U-shaped Transformer-based architecture designed for efficient and effective SIRR.

    \item We introduce a novel FFT Transformer block to replace standard self-attention. In this block, the spatial-domain branch extracts local features, while the frequency-domain branch captures global relationships.

    \item We present a simple yet effective HiT block, transforming popular Transformer-based methods into hierarchical Transformers. This improves SIRR performance by leveraging multi-scale features and long-range dependencies.
    
\end{itemize}

\section{Related Work}
\label{sec:rel}

Over recent decades, numerous innovative methods have been proposed to address the problem of reflection removal. Some methods rely on additional inputs, such as multi-frames~\cite{niklaus2021learned}, polarization~\cite{lei2020polarized}, and flash-only priors~\cite{lei2023robust}. Others utilize traditional image priors, including relative smoothness, ``ghosting'' cues, and dark channels, to separate and remove reflection artifacts. In contrast, our research is primarily dedicated to exploring and advancing deep learning techniques for effectively removing reflections from single images.

With the advancement of deep learning, learning-based SIRR methods~\cite{yang2025openrr1k,yang2025ntire,yang2025survey,cai2025openrr5k,cai2025vlm,dong2021location,song2023robust,li2023two,hu2023single,zhu2024revisiting,zhong2024language} have achieved remarkable performance gains across diverse reflection scenarios, becoming the leading solutions in this field.
This paper~\cite{dong2021location} presents a LANet for SIRR. It employs a reflection detection module that generates a probabilistic confidence map using multi-scale Laplacian features. The network, designed as a recurrent model, progressively refines reflection removal, with Laplacian kernel parameters highlighting strong reflection boundaries to improve detection and enhance the quality of the results.
V-DESIRR~\cite{prasad2021v} introduces a lightweight model for reflection removal using an innovative scale-space architecture, which processes the corrupted image in two stages: a Low Scale Sub-network (LSSNet) for the lowest scale and a Progressive Inference (PI) stage for higher scales. To minimize computational complexity, the PI stage sub-networks are significantly shallower than LSSNet, and weight sharing across scales enables the model to generalize to high resolutions without the need for retraining.
YTMT~\cite{hu2021trash} introduces a simple yet effective interactive strategy called "Your Trash is My Treasure". This approach constructs dual-stream decomposition networks by facilitating block-wise communication between the streams and transferring deactivated ReLU information from one stream to the other, leveraging the additive property of the components.
RobustSIRR~\cite{song2023robust} introduces a transformer-based model incorporating cross-scale attention, multi-scale fusion, and adversarial training for enhanced SIRR. 
RAG~\cite{li2023two} refines transmission layer estimation by leveraging predicted reflections. DSRNet~\cite{hu2023single} consists of two cascaded stages and a learnable residue module (LRM), where the first stage extracts hierarchical semantics and the second stage refines decomposition by addressing the linearity assumption. The framework in~\cite{zhu2024revisiting} comprises RDNet and RRNet, with RDNet employing a pretrained backbone and interpolation for reflection mask estimation, and RRNet utilizing this estimate for reflection removal. A language-guided approach~\cite{zhong2024language} aligns textual descriptions with image layers through cross-attention and contrastive learning, aided by a gated network and randomized training to mitigate layer ambiguity.
Compared to existing methods, our model exhibits superior generalization and robustness, effectively handling diverse reflection cases while maintaining high computational efficiency.

\section{Methodology}
\label{sec:meth}

\begin{figure}[!t]
\centering
\includegraphics[width=0.5\textwidth]{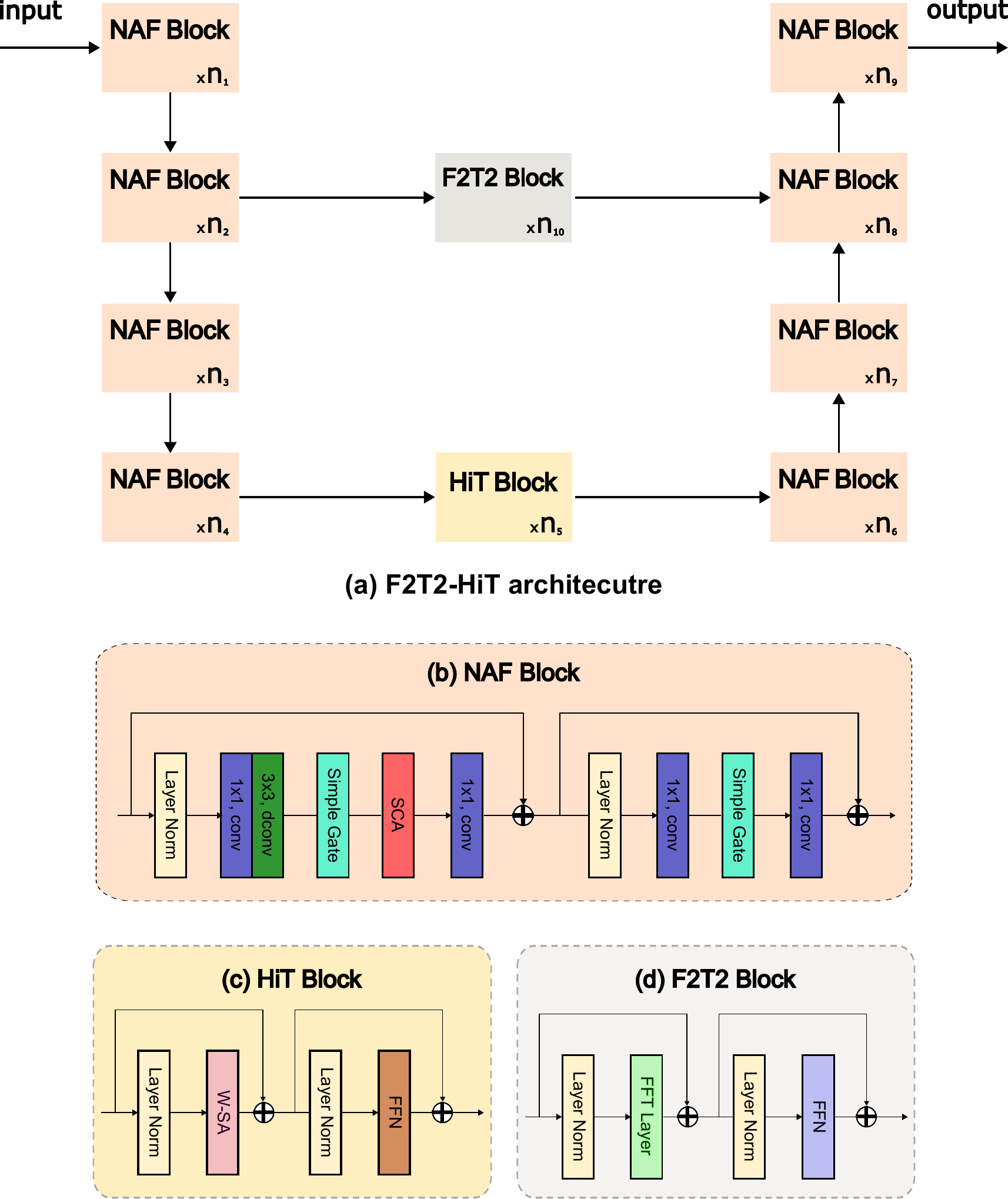}
\caption{An illustration of our U-shaped F2T2-HiT architecture. Certain details have been deliberately omitted for simplicity, such as downsampling/upsampling layers, feature fusion modules, input/output shortcuts, and more. Both (c) the HiT block and (d) the F2T2 block follow the structure of LayerNorm – Self-Attention (SA) – LayerNorm – Feed-Forward Network (FFN). Best viewed in color.}
\label{fig:UNet}
\end{figure}

We first introduce the fundamental methodology of the F2T2-HiT architecture in Sec.\ref{sec:F2T2-HiT}, followed by the block-level designs detailed in Sec.\ref{sec:NAFNet_Block}, Sec.\ref{sec:HiT_Transformer}, and Sec.\ref{sec:FFT_Transformer}, respectively.

\subsection{A U-Shaped F2T2-HiT Architecture}\label{sec:F2T2-HiT}

To simplify inter-block complexity, we utilize a classic single-stage U-shaped architecture with skip connections, as shown in Figure~\ref{fig:UNet}(a). 
For the plain block within the UNet structure, we incorporate three types of basic blocks: (1) the Nonlinear Activation Free (NAF) Block~\cite{chen2022simple}, shown in Figure~\ref{fig:UNet}(b); (2) the improved Hierarchical Transformer (HiT) Block~\cite{zhang2024hit}, shown in Figure~\ref{fig:UNet}(c); (3) the improved Fast Fourier Transform meets Transformer (F2T2) Block~\cite{jiang2024fast}, shown in Figure~\ref{fig:UNet}(d).

To maintain simplicity in the structure, we follow the principle of avoiding unnecessary additions to the architecture.
First, we apply Layer Normalization to all the plain blocks mentioned above, as it effectively stabilizes the training process and enhances performance, without incorporating other normalization techniques. 
Second, recent state-of-the-art (SOTA) methods increasingly favor replacing ReLU with GELU due to its smoother activation and improved gradient flow. We have also implemented this replacement in our model to enhance training stability and performance.
Third, depthwise convolution and channel attention have proven to be highly effective and efficient foundational modules. With their ability to reduce computational complexity and enhance global feature map fusion, they serve as essential elements in the design of our plain blocks.
Fourth, the standard self-attention mechanism~\cite{vaswani2017attention} generates target features through a similarity-weighted linear combination of all input features, allowing each feature to capture global information. Consequently, this attention mechanism plays a crucial role in building the internal structure of the block.
Finally, Fast Fourier Convolutions (FFCs)~\cite{chi2020fast} allow the incorporation of global context in the early layers, with a receptive field that spans the entire image. As a fully differentiable and efficient alternative to conventional convolutions, FFCs enhance high-resolution image restoration by utilizing global context early on and optimizing parameter usage for feature extraction and reconstruction.
The modules mentioned above have been empirically demonstrated to be highly effective for computer vision tasks. We construct our reflection removal model on these foundational components, which will further discuss in the following subsections.

\subsection{NAF Block}\label{sec:NAFNet_Block}

The NAFNet demonstrates exceptional performance in image restoration tasks like denoising and deblurring by utilizing the Nonlinear Activation Free (NAF) Block. This block integrates Layer Normalization for stable convergence, depthwise convolution for computational efficiency, simplified channel attention (SCA) to emphasize critical features, and gated linear units (GLUs) for dynamic feature control. By eliminating nonlinear activation functions, NAFNet simplifies the structure while maintaining strong feature representation and achieving a balance of efficiency and effectiveness in image restoration. We verified that the NAF Block can also achieve near state-of-the-art performance in the reflection removal task, this is why we adopted it as the foundational structure.

\subsection{Hierarchical Transformer Block}\label{sec:HiT_Transformer}

\begin{figure}[!t]
\centering
\includegraphics[width=0.5\textwidth]{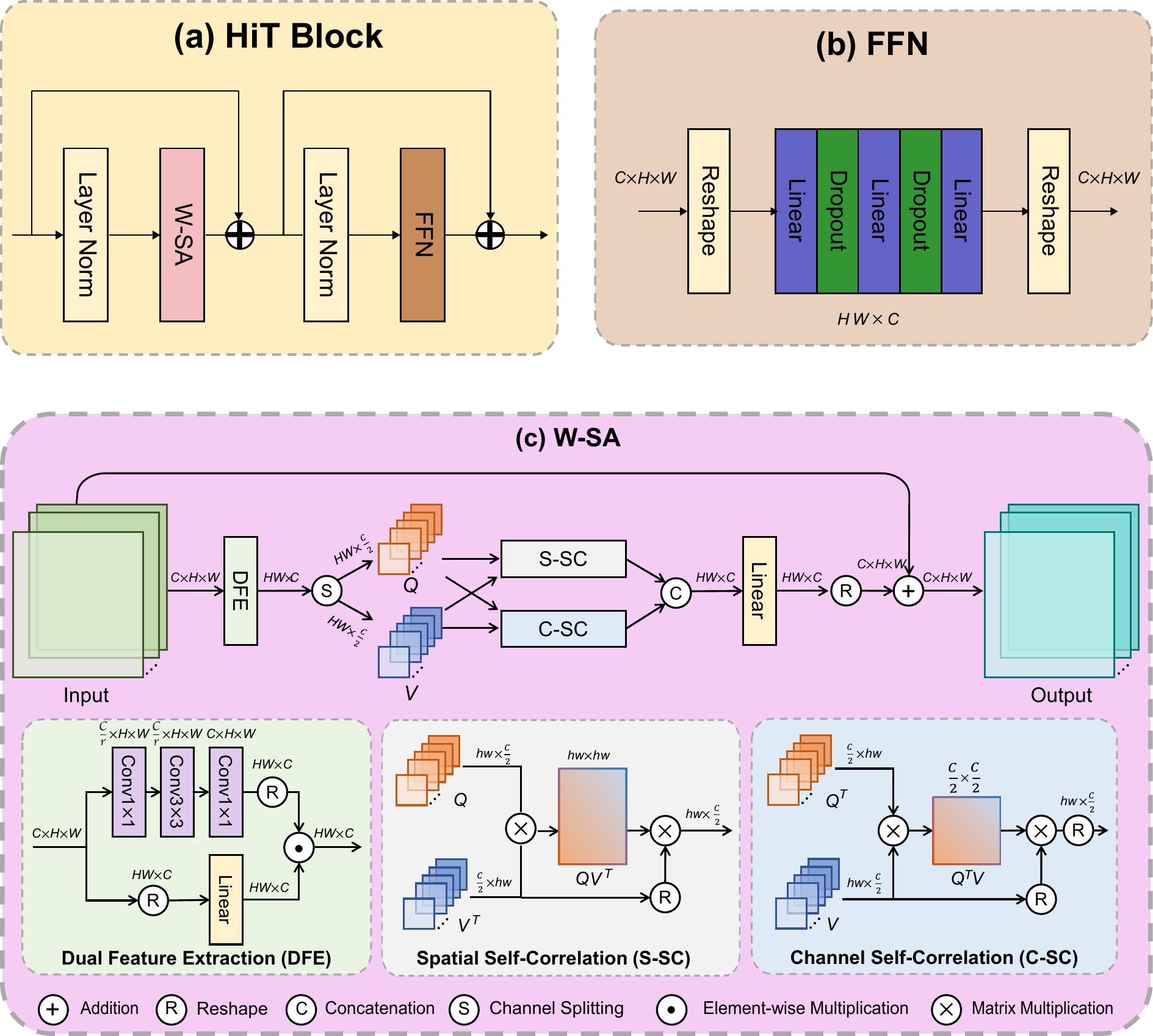}
\caption{An illustration of the proposed HiT Block. The channel-wise FFN is depicted in (b), which enhances the feature representation across channels. Window-based Self-Attention (W-SA) is shown in (c), consisting of DFE, S-SC, and C-SC. DFE is designed to extract features from both spatial and channel domains. S-SC and C-SC are proposed to efficiently aggregate hierarchical information with linear computational complexity relative to the window size. Queries (Q) and keys (K) are partitioned into non-overlapping windows based on the assigned window size. Best viewed in color.}
\label{fig:HiT}
\end{figure}

The standard self-attention mechanism~\cite{vaswani2017attention} generates the target feature as a weighted linear combination of all input features, where the weights are determined by the pairwise similarity between the features.
As a result, each feature captures global information but incurs quadratic computational complexity concerning the size of the feature map.
Some image restoration tasks involve processing high-resolution input, making the standard self-attention mechanism impractical.
Alternatively, ~\cite{liang2021swinir,liu2021swin} restrict self-attention to fixed-sized local windows to reduce computational costs, though this approach compromises global information.
Instead of using fixed window-based attention, we leverage multi-scale window sizes to effectively capture local information and global information. Specifically, as illustrated in Figure~\ref{fig:HiT}, we replace the commonly used fixed small windows with expanding hierarchical windows of sizes 4, 8, and 16. This design facilitates feature aggregation across multiple scales and enables the establishment of long-range dependencies. 
To further address the computational cost of large windows, we utilize a spatial-channel correlation method with linear complexity relative to window size, efficiently aggregating spatial and channel information from hierarchical windows.

\subsection{FFT Transformer Block}\label{sec:FFT_Transformer}

\begin{figure}[!t]
\centering
\includegraphics[width=0.5\textwidth]{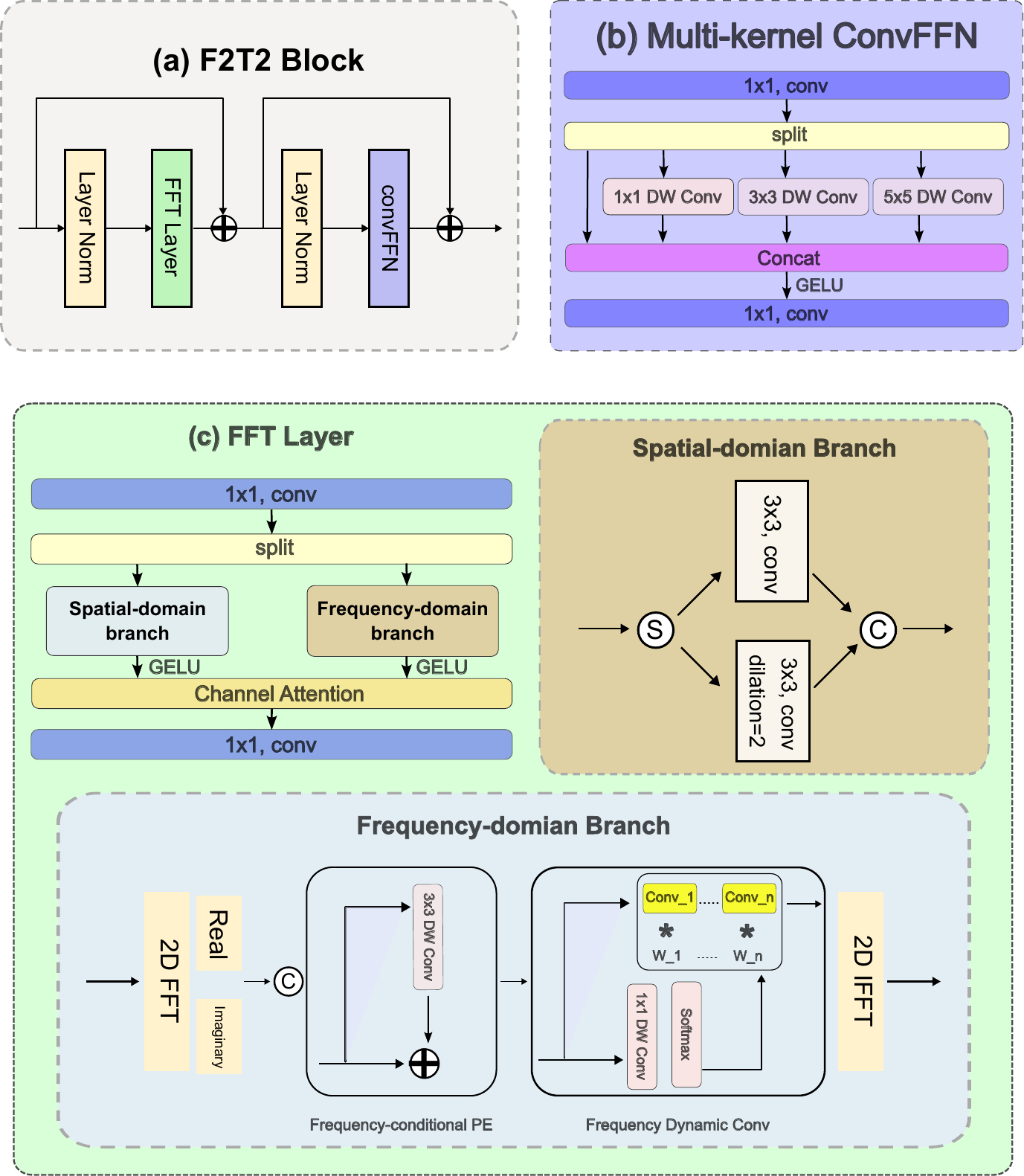}
\caption{An illustration of the proposed F2T2 Block. The spatial-wise FFN is depicted in (b), using parallel multi-kernel depthwise convolutions to capture spatial features at various scales. FFT layer is shown in (c), consisting of spatial-domain branch and frequency-domain branch. Best viewed in color.}
\label{fig:F2T2}
\end{figure}

Modern image reflection removal systems have made significant progress but still face challenges in completely removing large-area reflections, handling complex reflections, and processing high-resolution images. We identify that a key limitation contributing to these issues is the lack of an effective receptive field in reflection removal networks.

To address this challenge, we use FFCs, which provide an image-wide receptive field for comprehensive feature extraction. As shown in Figure~\ref{fig:F2T2}, we propose the F2T2 block, which integrates the FFT mechanism into the Transformer architecture. This design enables a dual-domain hybrid structure for multi-scale receptive field modeling. The spatial domain handles local feature extraction, while the frequency domain focuses on global modeling. By combining these two domains, our method effectively captures both local and global dependencies, enhancing performance in reflection removal task.

\begin{table*}[ht]
\centering
\caption{Quantitative comparisons on the real reflection benchmarks. The top-performing results are highlighted in \textbf{bold}, while the second-best results are marked with \underline{underlines}.
}
\label{tab:performance}
\begin{tabular}{crcccccccc}
\toprule
\multirow{2}{*}{Methods}   & \multirow{2}{*}{Venue} & \multicolumn{2}{c}{$Nature$ (20)} & \multicolumn{2}{c}{$Real$(20)} & \multicolumn{2}{c}{$SIR^{2}$(454)} & \multicolumn{2}{c}{$Average$(494)} \\  \cmidrule(l{2pt}r{2pt}){3-4} \cmidrule(l{2pt}r{2pt}){5-6} \cmidrule(l{2pt}r{2pt}){7-8} \cmidrule(l{2pt}r{2pt}){9-10}
&   & PSNR$\uparrow$         & SSIM$\uparrow$        & PSNR$\uparrow$        & SSIM$\uparrow$       & PSNR$\uparrow$       & SSIM$\uparrow$       & PSNR$\uparrow$     & SSIM$\uparrow$         \\ \hline
Input Image      & -    & 20.44     & 0.785   & 18.96    & 0.733   & 22.76   & 0.885   & 22.51   & 0.878  \\ \hline
BDN~\cite{yang2018seeing}  &  ECCV 2018 &   18.83 &  0.738  &  18.64   &  0.726   &  21.61  &   0.854   &   21.50    &    0.844          \\
RMNet~\cite{wen2019single} &  CVPR 2019 &   21.08 &  0.730 &  19.93 &  0.718  &   21.66  &   0.843   &  21.57   &  0.834            \\
ERRNet~\cite{wei2019single}   &  CVPR 2019     &  22.57     &    0.807    &   20.67    &    0.781   &   22.97    &   0.885    &   22.85  &   0.877     \\
Kim et al.~\cite{kim2020single} & CVPR 2020   &   20.10   &    0.759  &  20.22  & 0.735  &   23.57 &    0.877   &   23.30   &    0.886      \\
IBCLN~\cite{li2020single} & CVPR 2020  &   23.90   &    0.787  &    21.42  & 0.769     &   24.05  &    0.888     &   23.94 &    0.878          \\
YTMT~\cite{hu2021trash}  &  NerIPS 2021 &   20.69  &    0.777   &   22.94  &   0.815    &  23.57   &  0.889    &    23.43    &   0.882           \\
LANet~\cite{dong2021location}  &   ICCV 2021   &  23.51   &   0.810   &  \underline{23.40}   &    \textbf{0.826}  &   23.04     &  0.898  &    23.07    &  0.891   \\
PNACR~\cite{wang2023personalized}    & ACM MM 2023   & 23.93  & 0.807 &   22.57  &    0.806  &   24.14  & 0.894    &   24.06    &   0.888           \\
DSRNet~\cite{hu2023single}   &    ICCV 2023 &  21.24 &  0.789   &   22.32    &   0.806  & 24.91    &  0.902     & 24.65    &  0.893    \\
Zhu et al.~\cite{zhu2024revisiting}   & CVPR 2024  & \underline{25.96} & \textbf{0.843}  & \textbf{23.82}  & \underline{0.817}   & \underline{25.45}  & \textbf{0.910} &  \underline{25.40}   & \textbf{0.904}  \\   \bottomrule  
\\
Ours  & -   & \textbf{26.08}  & \underline{0.837}  & 21.64  & 0.766  & \textbf{25.72}  & \underline{0.903}  &  \textbf{25.57}  & \underline{0.894}  \\   \bottomrule   
\end{tabular}
\end{table*}

\begin{table*}[ht]
\centering
\caption{Ablation studies on the real reflection benchmarks. The best results are in \textbf{bold}, and the second-best results are \underline{underlined}.}
\label{tab:ablation_study}
\begin{tabular}{crcccccccc}
\toprule
\multirow{2}{*}{Methods}  & \multicolumn{2}{c}{$Nature$(20)} & \multicolumn{2}{c}{$Real$(20)} & \multicolumn{2}{c}{$SIR^{2}$(454)} & \multicolumn{2}{c}{$Average$(494)} \\  \cmidrule(l{2pt}r{2pt}){2-3} \cmidrule(l{2pt}r{2pt}){4-5} \cmidrule(l{2pt}r{2pt}){6-7} \cmidrule(l{2pt}r{2pt}){8-9}
& PSNR$\uparrow$   & SSIM$\uparrow$    & PSNR$\uparrow$    & SSIM$\uparrow$   & PSNR$\uparrow$   & SSIM$\uparrow$  & PSNR$\uparrow$  & SSIM$\uparrow$  \\ \hline
Input Image       & 20.44   &  0.785   &  18.96  &  0.733  & 22.76    & 0.885   & 22.51  & 0.878   \\ \hline
NAFNet            &  24.09  &  0.812   &  20.74  &  0.748  & 24.11    & 0.891   & 23.97  & 0.882    \\
NAFNet+Restomer   &  24.37  &  0.821   &  20.97  &  0.759  & 24.97    & 0.897   & 24.78  & 0.888    \\
NAFNet+HiT        &  \underline{25.51}  &  \underline{0.829}   &  \underline{21.16}  &  \underline{0.750}  & \underline{25.14}    & \underline{0.900}   & \underline{25.00}  & \underline{0.891}  \\  \bottomrule  \\
NAFNet+HiT+F2T2   & \textbf{26.08}  & \textbf{0.837}  & \textbf{21.64}  & \textbf{0.766}  & \textbf{25.72}  & \textbf{0.903}  &  \textbf{25.57}  & \textbf{0.894}    \\
\bottomrule
\end{tabular}
\end{table*}

\section{Experiments}
\label{sec:expe}

\subsection{Implementation details}\label{sec:Implementation_details}

Our framework is implemented with the PyTorch platform. During the training phase, the network is trained using the Adam optimizer with an initial learning rate of 0.0001, which is adjusted based on a Cosine Annealing Restart scheme. The scheduler is configured with three periods of 100,000 iterations each and corresponding restart weights of 1, 0.5, and 0.25. The total number of iterations is set to 300,000. The training is conducted with eight Nvidia A100 GPUs for approximately 24 hours. The batch size per GPU is set to 1, and 512 × 512 patches are randomly cropped from the images at each training iteration. Data augmentation includes random horizontal flipping and random rotation.

\subsection{Dataset and Evaluation Metrics}\label{sec:Dataset_Metrics}

Our training dataset includes both real and synthesized images. Following the approach~\cite{zhu2024revisiting}, the dataset comprises 89 real pairs from~\cite{zhang2018single}, 200 real pairs from ~\cite{li2020single}, synthesized pairs from the PASCAL VOC 2007 (5,011 images), PASCAL VOC 2012 (17,125 images)~\cite{everingham2010pascal}, and RRW dataset~\cite{zhu2024revisiting}. 
The synthesized image pairs are generated using the method outlined in~\cite{hu2023single}. 
For the testing dataset, consistent with previous methods, we evaluate the performance of our model by applying a single trained reflection removal model to three real-world reflection benchmarks: $Real$, $Nature$, and $SIR^{2}$. 
These three benchmarks, developed in different studies, encompass a wide range of real-world reflection scenarios and are commonly used to evaluate the performance of models in practical reflection removal tasks. However, it is important to note that among these datasets, $Real$ and $Nature$ include both training and test data, while $SIR^{2}$ is specifically designed as a test dataset.

As shown in Table~\ref{tab:performance}, our method outperforms most of the competitors across all testing datasets. Given that the three real-world datasets feature a wide range of scenes, lighting conditions, and glass thicknesses, it is a challenging task to achieve top performance across all metrics on these datasets simultaneously. The experimental results demonstrate that our proposed F2T2-HiT offers significantly higher performance and better generalization ability, highlighting the main contributions of our approach.

\subsection{Ablation Study}\label{sec:Ablation_Study}

\textbf{Impact of Hierarchical Transformer Block.}
To evaluate the effectiveness of the HiT Block, we conduct experiments to analyze its improvement over the Restormer Block~\cite{zamir2022restormer}, which incorporates standard Transformer components, such as multi-Dconv head transposed attention (MDTA) and Gated-Dconv feed-forward network (GDFN). However, Restormer has some limitations, including a high computational cost due to its complex attention mechanisms and the inefficiency in handling long-range dependencies, which our HiT Block aims to address through more efficient designs and better scalability.

\textbf{Impact of FFT Transformer Block.}
To evaluate the effectiveness of the FFT Transformer Block, we conduct experiments to analyze the improvements achieved by incorporating the FFT Transformer Block into the first skip-connection layers, comparing the results with those of a baseline model that does not use the F2T2 block. 

As shown in Table~\ref{tab:ablation_study}, the proposed HiT Block and F2T2 Block consistently surpass other baseline models on most testing datasets, underscoring the effectiveness of their design.

\section{Conclusion}
\label{sec:conclu}

In this paper, we propose F2T2-HIT, an innovative and efficient U-shaped Transformer-based architecture especially designed for SIRR task. By integrating two key innovations, we demonstrate significant improvements in reflection removal performance. First, the inclusion of HiT blocks allows the model to capture long-range dependencies and global image structures across multiple scales, which is crucial for effectively handling complex reflections and diverse reflection patterns. Second, the incorporation of F2T2 blocks enables the model to efficiently extract global information, enhancing its ability to capture large-scale image features necessary for accurate reflection removal.

Extensive experiments have validated the effectiveness and efficiency of F2T2-HIT, achieving state-of-the-art results on challenging reflection removal benchmarks. We also conducted an ablation study, which further confirms the effectiveness of both the HiT and FFT blocks in improving performance. Our approach offers a promising solution to the reflection removal problem, paving the way for more robust and accurate image restoration techniques.

\vfill\pagebreak



\clearpage
{\small
\bibliographystyle{IEEEbib}
\bibliography{refs}
}

\end{document}